\documentclass[letterpaper]{article} % DO NOT CHANGE THIS
\usepackage{aaai25}  % DO NOT CHANGE THIS
\usepackage{times}  % DO NOT CHANGE THIS
\usepackage{helvet}  % DO NOT CHANGE THIS
\usepackage{courier}  % DO NOT CHANGE THIS
\usepackage[hyphens]{url}  % DO NOT CHANGE THIS
\usepackage{graphicx} % DO NOT CHANGE THIS
\usepackage{natbib}  % DO NOT CHANGE THIS AND DO NOT ADD ANY OPTIONS TO IT
\usepackage{multirow}
\usepackage{caption} % DO NOT CHANGE THIS AND DO NOT ADD ANY OPTIONS TO IT
\usepackage{algorithm}
\usepackage{algorithmic}
\usepackage{xcolor}
\usepackage{booktabs}

\usepackage{amsmath} % For mathematical symbols and environments
\usepackage{amsfonts} 
\usepackage{colortbl}
\usepackage{newfloat}
\usepackage{listings}
\DeclareCaptionStyle{ruled}{labelfont=normalfont,labelsep=colon,strut=off} % DO NOT CHANGE THIS
\floatstyle{ruled}
\newfloat{listing}{tb}{lst}{}
\floatname{listing}{Listing}
\title{When Video Coding Meets Multimodal Large Language Models: A Unified Paradigm for Video Coding}
\author{
    Pingping Zhang\textsuperscript{\rm 1},
    Jinlong Li\textsuperscript{\rm 2},
    Kecheng Chen\textsuperscript{\rm 1},
    Meng Wang\textsuperscript{\rm 3},
    Long Xu\textsuperscript{\rm 4},
    Haoliang Li\textsuperscript{\rm 1},
    Nicu Sebe\textsuperscript{\rm 2}, 
    Sam Kwong\textsuperscript{\rm 3}, 
    Shiqi Wang\textsuperscript{\rm 1}
}
\affiliations{
    \textsuperscript{\rm 1}City University of Hong Kong, 
    \textsuperscript{\rm 2}University of Trento,
    \textsuperscript{\rm 3}Lingnan University,
    \textsuperscript{\rm 3}Ningbo University
}

\usepackage{bibentry}
\newcommand{\jinlong}[1]

\begin{document}

\maketitle

\begin{abstract}
    Traditional video compression methods perform well at high bitrates but struggle to preserve fine-grained semantic information at low bitrates. Recently, with the blossom of Multimodal Large Language Models (MLLMs), Cross-modal compression techniques offer prospective solutions for improving video compression under low-bitrate conditions.
    In this paper, we propose a unified Cross-Modality Video Coding (CMVC) framework that integrates multimodal representations and video generative models. The encoder disentangles video into spatial and temporal components, which are mapped to compact cross-modal representations using MLLMs. 
    During decoding, different encoding-decoding modes are employed to acquire various video reconstruction quality, including Text-Text-to-Video (TT2V) for semantic preservation and Image-Text-to-Video (IT2V) for perceptual consistency.
    Additionally, we elaborate an efficient frame interpolation model using Low-Rank Adaptation (\textit{LoRA}) to improve the perceptual quality.
    Experiments show that TT2V achieves effective semantic reconstruction, while IT2V ensures competitive perceptual consistency. These findings suggest the potential of leveraging multimodal priors to improve video compression, offering promising future research directions.

\end{abstract}

\section{Introduction}
Traditional video compression~\cite{bhaskaran1997image,Bross2Overview,ma2019image} has primarily focused on signal-level reconstruction, optimizing components such as pixel intensities and motion vectors during encoding and decoding. This approach has yielded substantial advancements in high-bitrate scenarios, exemplified by codecs like VVC~\cite{Bross2Overview} and DCVC~\cite{li2021deep}, which deliver remarkable performance. However, these conventional methods often encounter limitations at low bitrates scenario, particularly when it comes to preserving fine-grained semantic information, crucial for applications such as disaster response and remote surveillance. In such contexts, reconstructed video is prone to the loss or blurring of key features.

To address these shortcomings at low bitrates, cross-modal compression (CMC) methods have emerged~\cite{li2021cross,zhang2023rethinking,Gao2024Cross}, which harness the complementary strengths of multiple modalities to enhance image representations under constrained bitrate conditions. Notable approaches, such as VR-CMC~\cite{li2021cross}, SCMC~\cite{zhang2023rethinking}, MCM~\cite{Gao2024Cross}, and CMC-Bench~\cite{li2024cmc}, exploit the transformation of images into textual formats (I2T) to generate semantically similar content. This transformation not only preserves crucial semantic information but also achieves significant reductions in storage requirements, resulting in higher compression ratios compared to conventional image data representations. However, the potential of cross-modal representations for video compression, particularly in maintaining both semantic and perceptual quality at low bitrates, remains to be relatively underexplored.

Unlike image compression, which interests primarily in spatial information, video compression additionally involves the representation of temporal dynamics, significantly increasing its complexity.
In this regard, Multimodal Large Language Models (MLLMs)~\cite{han2022show,ruan2023mm} offer distinct advantages due to their inherent capacity to analyze and interpret temporal relationships within video content. Recent researches~\cite{lin2023video,zhang2023video,fei2024video} have demonstrated that MLLMs excel at processing sequential data and capturing dependencies across temporal events. By leveraging these capabilities, MLLMs can generate compact and semantically rich textual representations of video content, facilitating efficient compression while preserving high-quality reconstruction, even in low-bitrate settings. As such, MLLMs present a promising paradigm for advancing video compression, offering an effective balance between encoding efficiency and semantic preservation.

\begin{figure*}
\centering\includegraphics[width=\linewidth]{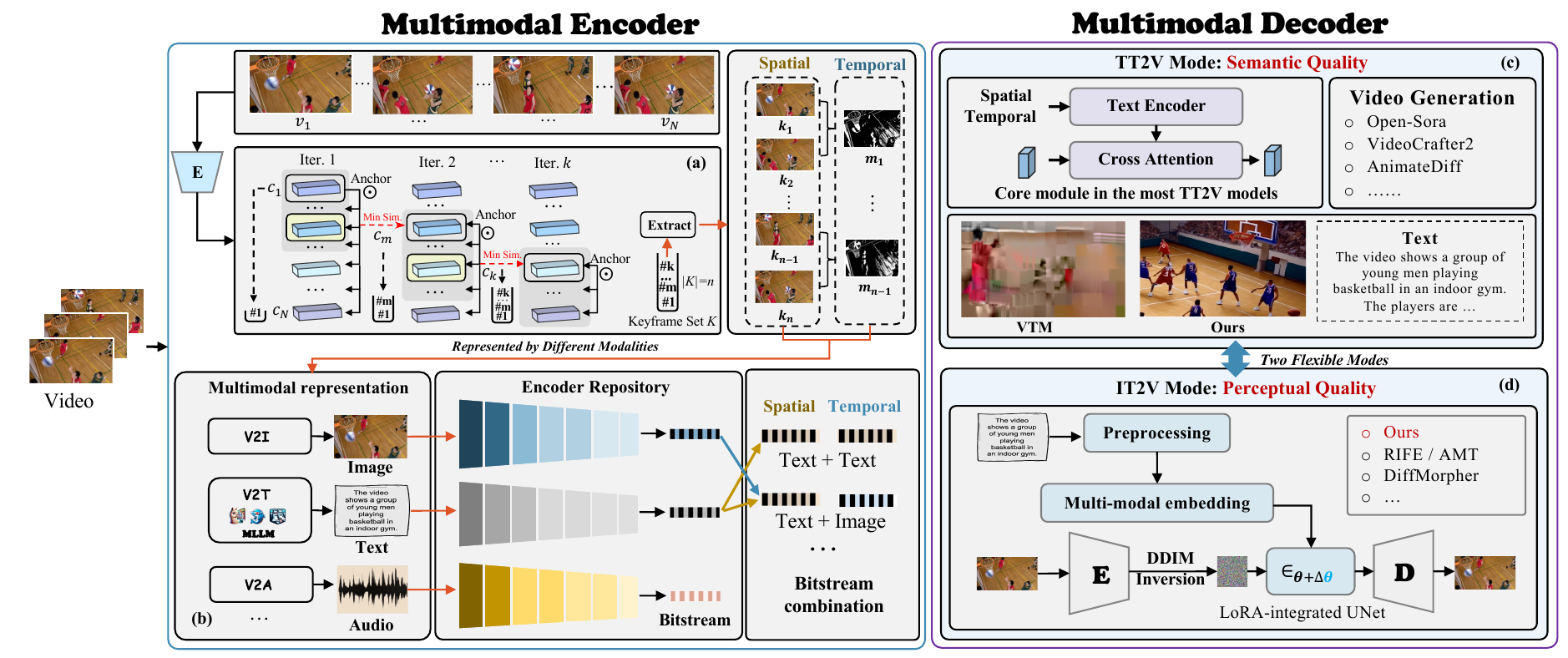}
    \caption{The framework of the proposed \textbf{CMVC} scheme. This framework operates by first segmenting the video into distinct clips using a keyframe selection strategy (a), allowing for the extraction of both spatial and temporal components from each video segment. Subsequently, MLLMs are employed to generate multimodal representations of these components. For instance, spatial information can be represented through text or images, while temporal dynamics may be encoded using text or audio modalities. These multimodal representations are then encoded via their respective encoders, resulting in compressed bitstreams for each component. The bitstreams corresponding to different components are then combined and transmitted to the decoder. In the decoder, we provide two exemplary modes, including TT2V (c) and IT2V (d) modes, for video generation. This model integrate various \textit{SoTA} models and mode conversions while maintaining semantic and perceptual quality at relatively high compression ratios.}
    \label{fig:framework}
    %\vspace{-0.3cm}
\end{figure*}

Motivated by the advantages of MLLMs, we propose a new Cross-Modality Video Coding (CMVC) scheme aimed at optimizing the representation of both spatial and temporal components, thereby enabling high-fidelity semantic and perceptual reconstruction at low bitrates. Capitalizing on the rich potential of multimodal representations, this framework supports the development of diverse encoding-decoding modes tailored to specific reconstruction requirements.
We propose two exemplary modes alternatively:
TT2V (Text-Text-to-Video) mode for semantic reconstruction at ultra-low bitrate (ULB) and IT2V (Image-Text-to-Video) mode for perceptual reconstruction at extremely low bitrate (ELB).
In the TT2V mode, inspired by the workflow of Cross-Modality Image Compression (CMIC)~\cite{li2021cross,zhang2023rethinking,li2024cmc} and the stunning generation capability of existing TT2V models for video reconstruction, we first extract the representative text constructed with our selection strategy, effectively encoding video content as the spatial component and motion as the temporal component. Then, the video generation model is utilized to reconstruct the corresponding video from text inputs. The rationale behind this strategy lies that compact yet effective text representations from the encoder encapsulate semantic details to enable high-quality semantic reconstruction for the decoder.
Different from TT2V, the IT2V mode is designed to enhance perceptual reconstruction, since images provide richer visual contexts compared to text, benefiting perceptual consistency. This is achieved by inputting similar text representations with the TT2V mode and extra selected keyframes from the encoder to the decoder for better perceptual video reconstruction. 
To further improve perceptual smoothness across consecutive frames, an efficient adaptation tuning in a frame-interpolation manner via Low-Rank Adaption (\textit{LoRA}) tuning is tailored to fully exploit the semantic cues and visual contexts from both input texts and keyframes to facilitate high-quality perceptual consistency for video reconstruction.
This comprehensive paradigm adeptly accommodates diverse modality representations within video coding, by tapping into foundational MLLMs and video generation models, which sheds light on future video coding works.
The contributions of our work are as follows:
\begin{itemize}
    \item To the best of our knowledge, the proposed unified paradigm for CMVC is the first to leverage foundational MLLMs and video generation models for video coding. 
    \item We elaborate multiple encoding-decoding modes to achieve good trade-off video reconstruction quality for specified decoding requirements, including TT2V mode to ensure high-quality semantic information and IT2V mode to achieve superb perceptual consistency. 
    \item Extensive experiments demonstrate that the proposed CMVC pipeline obtains competitive video reconstructions on HEVC Class B, C, D, E, UVG and MCL-JCV benchmarks while maintaining high compression ratios. 
\end{itemize}

\section{Related works}

\subsection{Video Generation Models} \label{video_gen_works}
Recently, video generation models have emerged as an increasingly promising topic, with numerous studies~\cite{ho2022imagen,singer2022make,ge2023preserve,blattmann2023align,chen2024videocrafter2,wang2023modelscope} showcasing promising advancements, that enables generative models simulate the real world principle. These include various approaches such as text-to-video (T2V)~\cite{wang2023modelscope,wang2023lavie}, image-to-video (I2V)~\cite{chen2023seine,esser2023structure,yin2023dragnuwa}, and IT2V~\cite{zhang2024diffmorpher}, among others.
T2V technology is designed to convert descriptive text into corresponding videos~\cite{lin2023video,wang2023modelscope,zhang2023video}. One of the primary challenges in this field is to understand the intricate semantics of the input text and effectively translate it into dynamic visual content, following real-world physics. To achieve optimal quality in the generated videos, these models are trained on large-scale video datasets, leveraging a large text-video corpus to train the model for better alignment.
I2V generative models typically include methods such as video interpolation and image-driven video diffusion. Image-driven video diffusion models~\cite{voleti2024sv3d,chai2023stablevideo,ouyang2024codef} necessitate the given referring image to steer the generative model to produce corresponding videos. Compared to image-driven video diffusion models, video interpolation techniques~\cite{huang2022rife,licvpr23amt} can better maintain consistency in both resolution and motions in terms of moving objects across consecutive frames, which elucidates great potential utility for image-driven video generation.
In contrast to I2V models, IT2V generative models incorporate textual guidance to enhance video generation. For instance, DiffMorpher~\cite{zhang2024diffmorpher} adds visual descriptions for images by adopting latent interpolation adaptation training to produce smoothing transformation. Building upon this concept, we can also enhance video generation by incorporating temporal descriptions.

\subsection{Cross-Modality Compression}
Multimodal generation has been effectively applied in the field of compression~\cite{lu2022learning}. To preserve semantic communication at ELB, CMC~\cite{li2021cross} integrates the I2T translation model with the T2I (Text-to-Image) generation model. Based on this foundation, SCMC~\cite{zhang2023rethinking} introduces a scalable cross-modality compression paradigm that hierarchically represents images across different modalities, thereby enhancing both semantic and signal-level fidelity.
Subsequently, VR-CMC~\cite{Gao2024Cross}, a variable-rate cross-modal compression technique, employs variable-rate prompts to capture data at varying levels of granularity. Additionally, a CMC benchmark has been established for image compression~\cite{li2024cmc}. These models demonstrate that the integration of I2T and T2I methodologies has outperformed the most advanced visual signal codecs.
Despite these advancements, there remains limited research focused on cross-modality video coding.

\section{CMVC Scheme}
\subsection{Overview}

We propose a CMVC paradigm for efficient video compression with high semantic and perceptual quality, especially at low bitrates, as illustrated in Fig.~\ref{fig:framework}. 
Given a video $V \in v_i$, where $v_i$ denotes video frames and $i \in \{1, \dots, N\}$, which consists of spatial (keyframe) and temporal (motion) components, the goal is to compress these components into compact multimodal representations.
We leverage MLLMs, specifically V2T models, to map both keyframes and motion into textual representations, which are then encoded using specialized encoders.
These multimodal representations are then compressed using dedicated encoders, yielding compressed representations of keyframe and motion.
The video is reconstructed by a decoder operating in one of two modes: TT2V, which prioritizes semantic consistency, and IT2V, which focuses on perceptual quality. 
This approach enables high compression ratios while preserving both semantic information and perceptual quality.
\subsection{CMVC Encoder}
\textbf{Keyframe selection strategy.}
Keyframes divide a full length video sequence into clips.
Let $n$ denote the number of keyframes, allowing us to extract $n$-1 clips from the video, with the first and last frames initially designated as keyframes. The first frame is encoded using the CLIP encoder~\cite{Radford2021LearningTV} to extract a high-level feature vector $v_{k}$ containing concise semantic information. We calculate the cosine similarity distance between the first frame and subsequent frames as follows:
\begin{equation}
    \centering
    \mathcal{D}_{c}=\frac{c_k \cdot c_{k+i}}{\left\|c_{k}\right\| \cdot\left\|c_{k+i}\right\|}=\frac{\sum_{j=1}^m c_{k, j} \cdot c_{k+i, j}}{\sqrt{\sum_{j=1}^m\left(c_{k, j}\right)^2 \cdot \sum_{j=1}^m\left(c_{k+i, j}\right)^2}},
    \label{eq.cosine}
\end{equation}
where $c_{k+i}$ is the feature vectors extracted from the subsequent frames. $m$ is the number of components of vectors $c_k$ and $c_{k+i}$. 
Within a uniform interval, we select the frame with the smallest similarity to the previous keyframes to form a set of keyframes that better showcase significant motion.
%to achieve a more compact and sparse keyframe samples.
Then, the $c_k$ is replaced by the next keyframe features, acting like a dynamic mechanism. This iterative process is repeated for subsequent clips, systematically identifying representative keyframes. 

\noindent\textbf{Multimodality representation.}
% disentangle video
In our proposed scheme, we focus on efficiently representing spatial and temporal information of videos through keyframes and motion.
Specifically, 
let $V$ represent the original video. The keyframes are denoted as $K$ = \{$k_1$, $k_2$, ..., $k_n$\}, where each $k_i$ is a keyframe. The motion information $m_j$ between two consecutive keyframes is represented by $M$ = \{$m_1$, $m_2$, ..., $m_{n-1}$\}.
Keyframes and motion are transformed into multimodality representations as follows: $T_{k, i} = f(k_i)$ and $T_{m, j} = g(m_j)$. 
Here, $f(*)$ and $g(*)$ denote the process of cross-modality representation for keyframes and motion, respectively. As illustrated in Fig.~\ref{fig:framework}, keyframes and motion can be transformed into textual and visual representations.
Thus, the total bitrate is given by: 
\begin{equation}
    R_{total} = \sum_{i=1}^{n} R_k(T_{k, i}) + \sum_{j=1}^{n-1} R_m(T_{m, j}),
\end{equation}
where $R_{k}$ and $R_{m}$ are the entropy coding modules for keyframes and motion, respectively. The bitrates can be adjusted by $n$ and the compression ratio of keyframes and motion.

\subsection{CMVC Decoder}
In the decoder, we utilize the decoded keyframe $\hat{K}$ and motion $\hat{M}$ to achieve video generation, as follows: 
\begin{equation}
    \hat{V} = \mathcal{G}(\hat{K}, \hat{M}),
\end{equation}
where $\mathcal{G}(*)$ is a video generation model and $\hat{V}$ is the reconstructed video.
Based on different modality representations for keyframe and motion, we designed two modes, including the TT2V mode and the IT2V mode for ULB and ELB coding, respectively. 

In the TT2V mode, we utilize state-of-the-art (\textit{SoTA}) video generation models to generate videos from decoded keyframes and motion descriptions. Leveraging advancements in these models, we optimize semantic reconstruction, with our results showing that more detailed descriptions yield higher bitrates and improved semantic quality. In the IT2V mode, keyframe images and motion descriptions are integrated to enhance perceptual quality. In addition to employing existing IT2V models, we propose a generative model utilizing \textit{LoRA} tuning to ensure superior perceptual consistency at ELB.

\noindent \textbf{The IT2V generative model.}
\begin{figure}
    \centering
    \includegraphics[width=\linewidth]{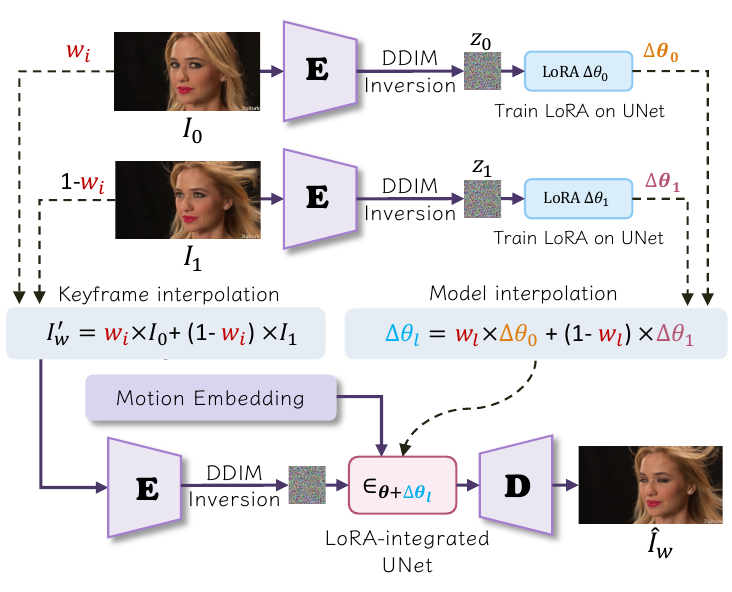}
    %\vspace{-0.3cm}
    \caption{The workflow of the IT2V generative model. Two LoRAs are trained to fit the two keyframe images ($I_{0}$ and $I_{1}$), respectively. To generate $w$-th frame between $I_0$ and $I_1$, we interpolate $I^{'}_{w}$ and the LoRA parameters according to the weights $w_i$ and $w_l$.}
    \label{fig:lora}
    \vspace{-0.10cm}
\end{figure}
%1. model framework
%
The IT2V mode is designed to obtain a reconstructed video according to keyframe images and the text of motion. Thus, we propose a IT2V generative model, which generates a video clip according to two keyframe images ($I_0$ and $I_1$) and the description of the motion of this video clip. Specifically, we adopt a stable diffusion model (SD) with \textit{LoRA}, which fine-tunes the model parameters $\theta$ by training a low-rank residual component $\triangle\theta$. This residual can be decomposed into products of low-rank matrices. LoRA demonstrates significant efficiency in generating various samples while maintaining consistent semantic identity across different latent noise traversals. The proposed IT2V model is shown in Fig.~\ref{fig:lora}. We first train two LoRAs ($\Delta \theta_0$ and $\Delta \theta_1$) on the SD UNet $\epsilon_\theta$ for each of the two images $I_0$ and $I_1$. The learning objective of $\Delta \theta_i(i=0,1)$ is: 
\begin{equation}
\mathcal{L}\left(\Delta \theta_i\right)=\mathbb{E}_{\epsilon, t}\left[\left\|\epsilon-\epsilon_{\theta+\Delta \theta_i}\left(\mathbf{z}_{t, i}, t, \mathbf{c}_i\right)\right\|^2\right],
\end{equation}
where $\mathbf{z}_{t, i}=\sqrt{\bar{\alpha}_t} \hat{\mathbf{z}}_{i}+\sqrt{1-\bar{\alpha}_t} \epsilon$ is the noised latent embedding at diffusion step $t$. $\hat{\mathbf{z}}_{i}$ is the VAE encoded latent of the $I_{i}$ image. $\epsilon \sim \mathcal{N}(\mathbf{0}, \mathbf{I})$ is the random sampled Gaussian noise. $\mathbf{c}_i$ is the motion embedding encoded from the motion prompt. $\epsilon_{\theta+\Delta \theta_i}$ represents the LoRA-integrated UNet. The fine-tuning objective is optimized separately via gradient descent in $\Delta \theta_0$ and $\Delta \theta_1$.

\begin{figure*}[t]
    \centering
    \includegraphics[width=1.0\linewidth]{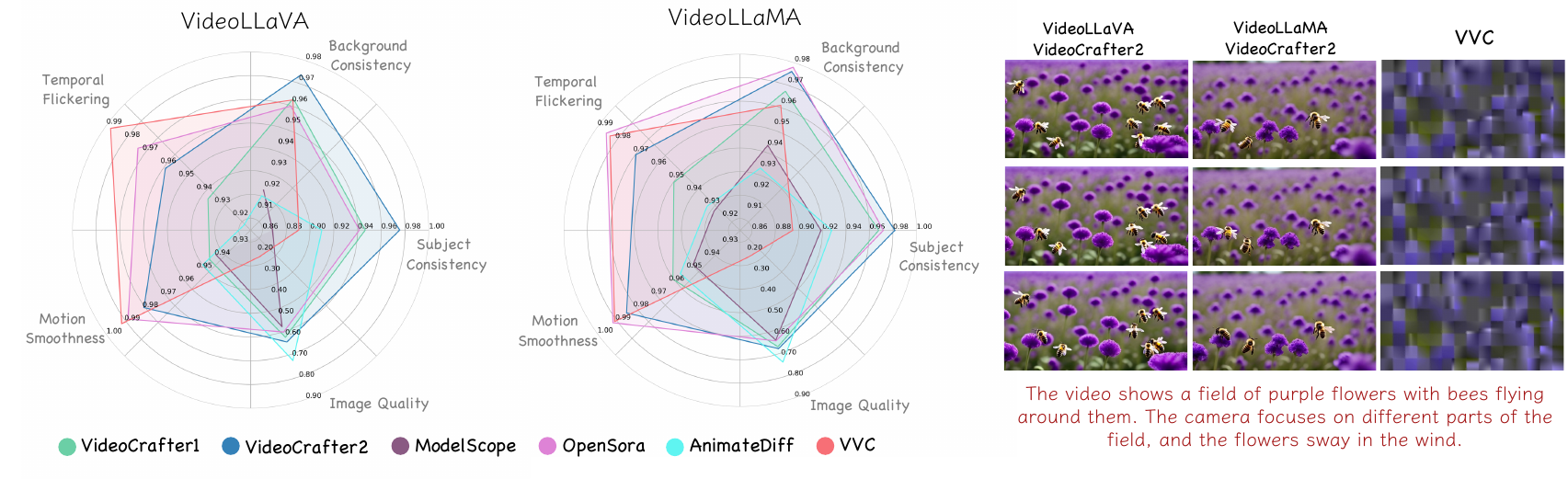}
    \caption{Left: Comparison results of combination of different V2T models (VideoLLaVA and VideoLLaMA) and TT2V models (VideoCrafter1, VideoCrafter2, ModelScope, OpenSora and AnimateDiff). Right: Visual quality comparison of the TT2V mode and VTM. At ULB, our proposed TT2V mode successfully preserves the semantic quality of the videos. In contrast, VTM brings significant blocking artifacts, which impedes the effective conveyance of semantic information in videos. }% The text was generated using the VideoLLaVA model.
    \label{fig:IT2V_results}
    \vspace{-0.1cm}
\end{figure*}

\begin{figure*}[t]
    \centering
    \includegraphics[width=\linewidth]{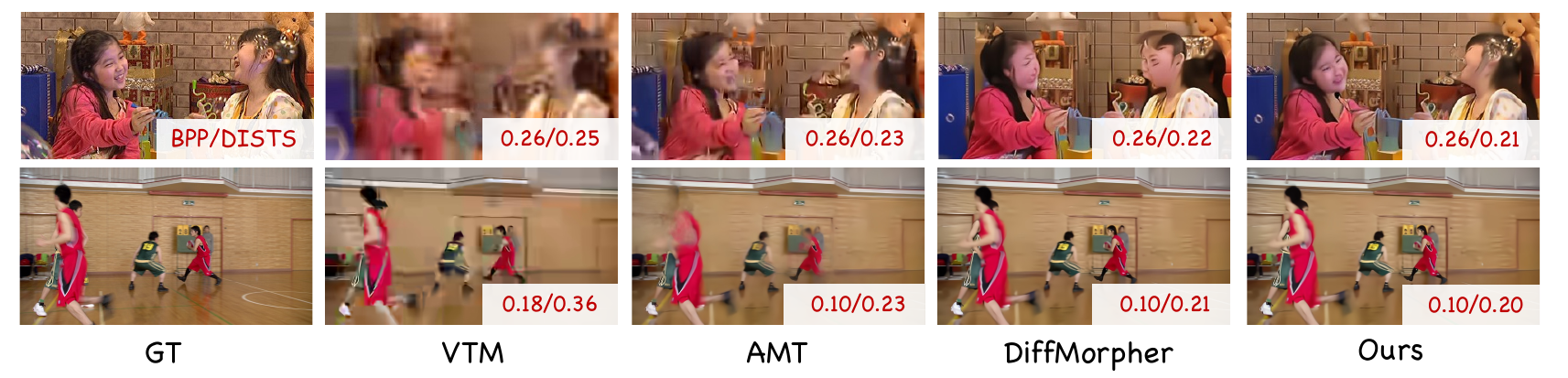}
    \caption{Visual quality comparison. The values represent the BPP (1e-2) and the DISTS value. A lower DISTS value indicates better perceptual quality.}
    \label{fig:IT2V_vis}
    \vspace{-0.1cm}
\end{figure*}

\noindent\textbf{Frame and model interpolation.}
In order to generate $I^{'}_w$, we first conduct keyframe interpolation as input in the following manner:
\begin{equation}
I_w^{\prime}=w_i \times I_0+\left(1-w_i\right) \times I_1.
\end{equation}
Building upon DiffMorpher~\cite{zhang2024diffmorpher}, we then interpolate the model weight $\Delta \theta_l$ according to $\Delta \theta_0$ and $\Delta \theta_1$:
\begin{equation}
    \Delta \theta_l=w_l \times \Delta \theta_0+\left(1-w_l\right) \times \Delta \theta_1.
\end{equation}
$\Delta \theta_l$ is the LoRA parameters, which are integrated to UNet $\epsilon_{\theta+\Delta \theta_l}$. 

The uniformly linear interpolation schedule may result in an uneven transition. Thus, we conduct online training for the $w_i$ and $w_l$ with the constraint of $D(I_w, \hat{I}_{w})$ at the encoder side, where $D(*)$ is the $\mathcal{L}_2$ loss. We only update the $w_i$ and $w_l$, as follows:
\begin{equation}
    w^{t+1}_{i} = w^{t}_{i} - \alpha \nabla D(w^{t}_{i}),
\end{equation}
\begin{equation}
    w^{t+1}_{l} = w^{t}_{l} - \alpha \nabla D(w^{t}_{l}),
\end{equation}
where $w^{t}_{i}$ and $w^{t}_{l}$ is the parameter $w_i$ and $w_l$ at training step $t$, resepctively. $\alpha$ is the learning rate, set to 0.001, while $\nabla D(*)$ denotes the gradient of the loss function concerning the parameters at the training step $t$. After obtaining optimal $w^{t}_{i}$ and $w^{t}_{l}$, we compress and transmit them to the decoder. The VAE decoder then reconstructs the denoised latent representation into the $w$-th frame, resulting in $\hat{I}_{w}$.

\section{Experiments}
\subsection{Experimental settings}

\noindent \textbf{Datasets.}
The datasets, including HEVC Class B, C, D, and E, as well as UVG and MCL-JCV, are extensively utilized for evaluating both traditional and neural video codecs. These datasets vary in resolution and content, providing a diverse range of scenarios for comprehensive assessment. To ensure compatibility with various video codecs, we resize videos to dimensions that are multiples of 64 for both width and height.

\begin{figure*}[t]
    \centering
    \includegraphics[width=\linewidth]{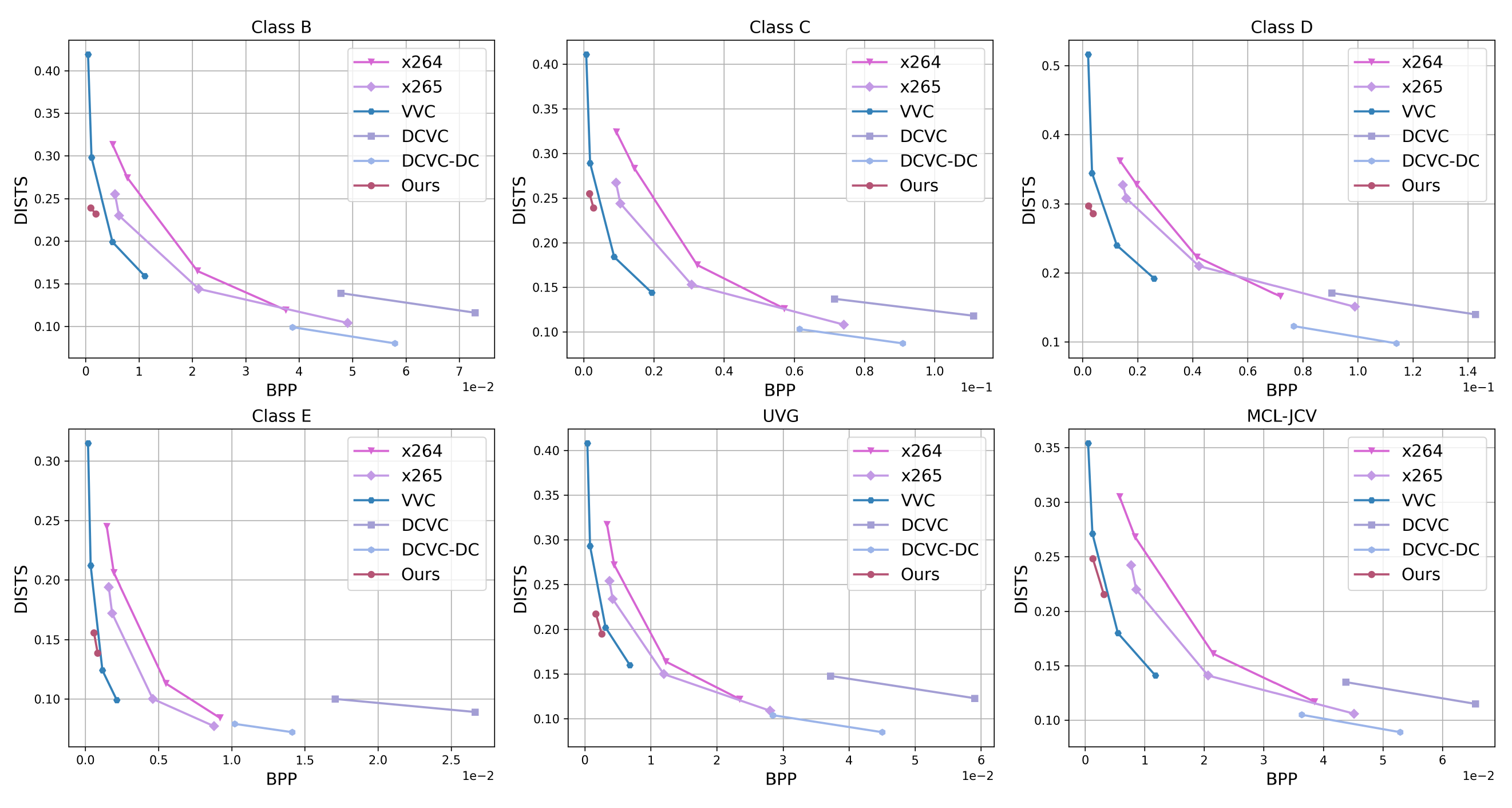}
    \caption{The R-D performance comparison in the IT2V mode. The comparisons are performed on the Class B, Class C, Class D, Class E, UVG, and MCL-JCV, respectively.}
    \label{fig:IT2V_DISTS}
    \vspace{-0.3cm}
\end{figure*}

\begin{table}[t]
    \centering
    \small
    \caption{BD-Rate (\%) comparison of different video generation models across various datasets in terms of DISTS. The anchor is VTM with QP=\{52, 50, 47, 45\}.}
    %\vspace{-0.05cm}
    \begin{tabular}{crrrr}
    \toprule
        \textbf{} & RIFE & AMT & DiffMorpher & Ours \\ 
    \toprule
        Class B & 4.32  & 53.97   &  -45.94    \quad\qquad & \textbf{-59.12}  \\ \midrule
        Class C & 42.58  &50.57   &  \textbf{-31.06}  \quad\qquad& -17.27  \\ \midrule
        Class D & -4.89 &-29.88  &  -49.06 \quad \qquad& \textbf{-52.16}  \\ \midrule
        UVG & 24.08      &15.00   &  11.55       \quad\qquad & \textbf{-21.18}  \\ \midrule
        MCL-JCV & 83.08  &151.20  &  22.16    \quad\qquad & \textbf{-7.25} \\ 
        \bottomrule
    \end{tabular}
    \vspace{-0.2cm}
    \label{tab:bdbr}
\end{table}

\noindent \textbf{Comparison methods.}
There are numerous SOTA foundation models available for video understanding. We choose two prominent models, namely VideoLLaVA~\cite{lin2023video} and VideoLLaMA~\cite{zhang2023video}, to extract semantic information from videos. This process aligns with the V2T stage depicted in Fig.~\ref{fig:framework}, where the selected models play an important role in extracting semantic descriptions for keyframes and motion.
In the TT2V mode, numerous video generation models are available. In this context, we employ advanced video generation models, including Open-Sora~\cite{pku_yuan_lab_and_tuzhan_ai_etc_2024_10948109}, VideoCrafter1~\cite{chen2023videocrafter1}, VideoCrafter2~\cite{chen2024videocrafter2}, and AnimateDiff~\cite{guo2023animatediff}, for the purpose of generating videos based on textual input. In addition, we compare with the video codec VTM at the extremely low bitrate with QP=63. 
In the IT2V mode, we conduct a comparative analysis of existing traditional video codecs, such as x264, x265, and VTM~\cite{Bross2Overview}. Alongside this, we evaluate our method against deep video codecs such as DCVC~\cite{li2021deep} and DCVC-DC~\cite{li2023neural}, but these codecs encounter challenges in achieving extremely low bitrate coding.
In addition, we compare the video generation technique, DiffMorpher~\cite{Zhang_2024_CVPR}, which requires keyframe images and motion descriptions for controlling video generation. In our exploration of various video interpolation methods, it is essential to note that these approaches rely solely on keyframes for control, omitting any incorporation of motion descriptions. Furthermore, it should be emphasized that the bit consumption associated with motion text has not been calculated. 
\begin{figure}[t]
    \centering
     \vspace{-0.1cm}
     \includegraphics[width=0.9\linewidth]{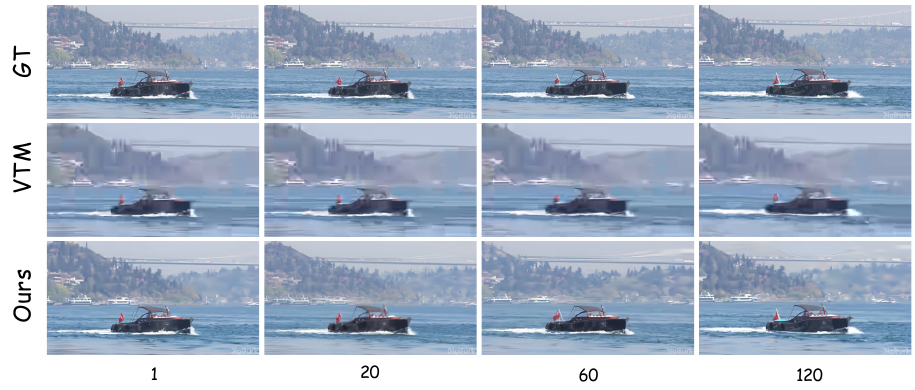}
    \caption{Visual quality comparison in a video. The numbers displayed beneath the images correspond to the frame index.}
    \label{fig:compare}
    \vspace{-0.3cm}
\end{figure}

\begin{figure}[t]
    \centering
    \vspace{-0.2cm}
    \includegraphics[width=0.98\linewidth]{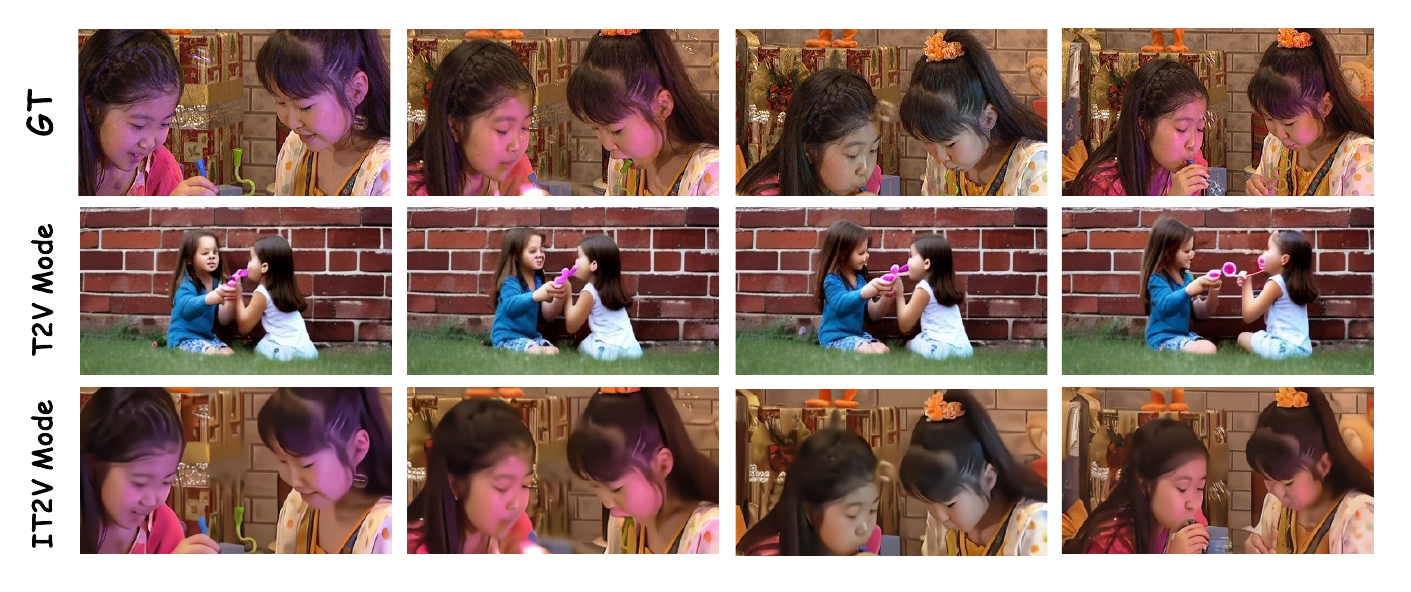}
    \caption{Visual quality comparison between the TT2V mode and the IT2V mode. The TT2V mode effectively preserves semantic consistency with the ground truth, while the IT2V mode is designed to keep perceptual consistency. }
    \label{fig:t2v_it2v_vis_compare}
    \vspace{-0.35cm}
\end{figure}

\subsection{Experimental results}
\noindent \textbf{Comparison in the TT2V mode.} We conduct a comparative analysis of two V2T models, VideoLLaVA and VideoLLaMA, both of which are the SOTA MLLMs. Subsequently, we extend our comparison to include five video generation models: VideoCrafter1, VideoCrafter2, ModelScope, OpenSora, and AnimateDiff. Furthermore, we compare these models against the traditional video codec VTM at QP=63, which results in a higher bitrate than our proposed scheme. Our assessment focuses on five aspects: subject consistency, background consistency, temporal flickering, motion smoothness, and frame quality. The results, illustrated in Fig.~\ref{fig:IT2V_results}, indicate that the TT2V generation models outperformed VTM, showcasing better frame-wise quality and consistency in both background and subject representation. These results reflect the average performance across all testing datasets, and detailed comparison results can be found in the supplementary material. The visual quality comparisons illustrated in Fig.~\ref{fig:IT2V_results} indicate that VTM displays considerable blocking artifacts, which severely hinder its ability to convey semantic information.

\begin{table}[t]
    \centering
    \small
    \caption{Ablation studies on the keyframe selection strategy.}
    \vspace{-0.1cm}
    \begin{tabular}{ccr}
    \toprule
         \textbf{Models} & Settings & BD-Rate(\%) $\downarrow$ \\ 
    \toprule
    % compared under the same quality and different sampling number
    \multirow{4}{*}{Sampling strategies}  & Uniform sampling  & -19.705 \quad 
    \quad \\
     & Random sampling  &  -9.859 \quad 
    \quad \\
     & MSE  & -14.496 \quad 
    \quad \\
     & CS  & -24.206  \quad 
    \quad \\
    \midrule

    \multirow{3}{*}{Keyframe number} &  2 &  -20.001\quad \quad \\ 
                                     &  3 &  -5.104  \quad \quad \\ 
                                     &  4 &  12.665  \quad \quad \\
                                     
    \midrule
    \multirow{3}{*}{Keyframe quality} &   low    &  -24.206 \quad \quad \\ 
                                      &   middle &  -12.043  \quad \quad \\ 
                                      &   high   &  -2.617  \quad \quad \\ 
        
    \bottomrule
    \end{tabular}
    \vspace{-0.25cm}
    \label{tab:ab_keyframe}
\end{table}

\noindent \textbf{Comparison in the IT2V mode.} 
We compare our model with traditional codecs (x264, x265, and VTM) as well as deep video codecs (DCVC and DCVC-DC). As presented in Fig.~\ref{fig:IT2V_DISTS}, we use DISTS to evaluate perceptual quality. Additional comparisons with other evaluation metrics, such as LPIPS, FID, and PSNR, are provided in the supplementary material. However, the pretrained models provided by deep video codecs have limitations in achieving ELB.
In addition, we compare our proposed model with various video generation models, including RIFE~\cite{huang2022rife}, AMT~\cite{licvpr23amt}, and DiffMorpher~\cite{zhang2024diffmorpher}, as detailed in Table~\ref{tab:bdbr}. By adjusting the number and quality of keyframe images, we can effectively control the bitrate. For our comparisons, we select the optimal results for comparison, where the settings can be found in the supplementary material. Our model exhibits superior performance across most datasets, demonstrating greater stability compared to other video generation models. The visual quality is evaluated at similar bitrates, as shown in Fig.~\ref{fig:IT2V_vis} and Fig.~\ref{fig:compare}. Our proposed model exhibits superior perceptual quality in both spatial and temporal dimensions. Additionally, we showcase frames sampled from the decoded videos generated by the TT2V mode and the IT2V mode. The TT2V mode effectively preserves semantic consistency with the ground truth, while the IT2V mode further ensures perceptual consistency.

\subsection{Ablation studies}
\noindent\textbf{Keyframe.} We perform ablation studies focused on keyframes, examining various aspects such as keyframe selection methods, the quality of keyframe images, and the number of keyframe images. In our keyframe selection process, we evaluate various sampling strategies, including uniform sampling and random sampling. Given that these techniques do not rely on a distance function, we also compare the sampling strategy with mean-square error (MSE) distance to Ours with cosine similarity (CS) distance. Regarding the quality of keyframe images, we varied the quality levels, including low, medium, and high quality, which correspond to the compression factors 64, 128, and 256, respectively. The results presented in Table~\ref{tab:ab_keyframe} indicate that higher quality decoded images result in increased bitrate consumption, such that higher quality does not necessarily lead to a better BD-Rate. Adjusting the number of keyframes based on the frame number of the video, we observe that a lower number of keyframes can maintain a balance between quality and bitrate consumption.

\noindent\textbf{IT2V generative model.} We perform ablation studies focused on the different settings, including the influence of motion description, different codecs, updating strategies, training step, and sampling step. In terms of motion description, we compare the model without motion description, as depicted in Table~\ref{tab:IT2V_ours}. The results indicate that incorporating motion description significantly enhances video reconstruction quality. Additionally, we explore a range of codecs for keyframe images, such as Hyperprior~\cite{balle2018variational}, NIC~\cite{Chen2021End}, and NVTC~\cite{feng2023nvtc}. Among these, NVTC stands out by demonstrating superior reconstructed quality while maintaining a lower coding rate.
Our model requires updating $w_i$ and $w_{l}$ based on the input, such that we further evaluate the effectiveness of updating strategies, as illustrated in Table~\ref{tab:IT2V_ours}. To assess the effectiveness of these updating strategies, we present further evaluations in Table~\ref{tab:IT2V_ours}. Moreover, we examine the repercussions of varying training and sampling steps. An increase in the number of sampling steps correlates with improved results.
To strike a balance between performance and computational efficiency, we choose 100 training steps and 50 sampling steps for our final implementation.

\begin{table}[t]
    \centering
    \small
    \caption{Ablation studies of the IT2V mode.}
    \vspace{-0.1cm}
    \begin{tabular}{ccr}
    \toprule
         \textbf{Models} & Settings & BD-Rate(\%) $\downarrow$ \\ 
    \toprule
    \multirow{2}{*}{Motion description}            &  $\times$    &  8.516 \quad \quad \\ 
     & $\checkmark$    &  -24.206 \quad \quad \\ \midrule
    
    \multirow{3}{*}{Codecs} &   Hyperprior & -12.261 \quad \quad \\ 
                            &   NIC & -16.132 \quad \quad \\ 
                            &   NTVP & -24.206 \quad \quad \\ 

   \midrule
    
    Updating $w_{i}$              &  $\times$    &  -12.402 \quad \quad \\
    Updating $w_{l}$              &  $\times$    &  -18.551 \quad \quad \\ 
    Updating $w_{i}$ and $w_{l}$  &  $\times$    &   -2.351 \quad \quad \\ 
    Updating $w_{i}$ and $w_{l}$  &  $\checkmark$    &   -24.206 \quad \quad \\\midrule

    \multirow{3}{*}{Training step}   &  50  &    -17.864 \quad \quad \\ 
                                     &  100 &    -24.206 \quad \quad \\ 
                                     &  150 & -25.196   \quad \quad \\\midrule
    \multirow{3}{*}{Sampling step}   &  20 &    -4.035 \quad \quad \\ 
                                     &  50 &   -24.206 \quad \quad \\  
                                     &  100 &  -25.207 \quad \quad \\

    \bottomrule
    \end{tabular}
    \vspace{-0.22cm}
    \label{tab:IT2V_ours}
\end{table}

\section{Discussion}
\noindent\textbf{Application.}
The proposed method enables efficient transmission at low bitrates while preserving semantic content, making it ideal for bandwidth-limited or emergency alert scenarios. When bandwidth allows, keyframe data can be transmitted with textual descriptions, allowing the decoder to reconstruct an approximate video. This hybrid approach balances visual quality with data efficiency, suitable for situations where full video streaming is infeasible, such as disaster response and remote surveillance. Although still in the research phase, the method shows strong potential for real-world applications. With advances in computational resources and model optimization techniques like pruning and quantization, it is expected to become a practical solution for emergency communications and other bandwidth-constrained environments. 
% These optimizations will enhance transmission efficiency, reduce computational load, and ensure the timely delivery of critical information in resource-limited scenarios.

\noindent\textbf{Higher bitrate.} CMVC includes the TT2V mode and IT2V mode, but more modes can be further explored. For instance, motion representation can be realized through optical flow or trajectories. By integrating multiple modalities of keyframes and motion, we can cater to diverse reconstruction requirements.
Furthermore, future efforts should prioritize enhancing CMVC at higher bitrates by integrating more control information to facilitate the reconstruction of the original video. This approach aims to achieve superior performance across all bitrates and dimensions when compared to traditional codecs.

\section{Conclusion}
We propose a CMVC paradigm that represents a promising advancement in video coding technology. This framework effectively tackles the challenges of preserving semantic integrity and perceptual consistency at ULB and ELB.
By leveraging MLLMs and cross-modality representation techniques, the proposed CMVC framework disentangles videos into content and motion components, transforming them into different modalities for efficient compression and reconstruction.
Through the TT2V and IT2V modes, CMVC achieves a balance between semantic information and perceptual quality, offering a comprehensive solution at high compression ratios.

\newpage
\bibliography{aaai25}

\end{document}